# Approximation Algorithms for the Loop Cutset Problem

**Ann Becker and Dan Geiger**
Computer Science Department
Technion
Haifa 32000, ISRAEL
anyuta@cs.technion.ac.il, dang@cs.technion.ac.il

## Abstract

We show how to find a small *loop cutset* in a Bayesian network. Finding such a loop cutset is the first step in the method of conditioning for inference. Our algorithm for finding a loop cutset, called MGA, finds a loop cutset which is guaranteed in the worst case to contain less than twice the number of variables contained in a minimum loop cutset. We test MGA on randomly generated graphs and find that the average ratio between the number of instances associated with the algorithms' output and the number of instances associated with a minimum solution is 1.22.

## 1 Introduction

Most inference algorithms for the computation of a posterior probability in general Bayesian networks have two conceptual phases. One phase handles operations on the graphical structure itself and the other performs probabilistic computations. For example, the clique tree algorithm requires us to first find a "good" clique tree and then perform probabilistic computations on the clique tree [LS88]. Pearl's method of conditioning requires us first to find a "good" loop cutset and then perform a calculation for each loop cutset [Pe86, Pe88]. Finally, Shachter's algorithm requires us to find a "good" sequence of transformations and then, for each transformation, to compute some conditional probability tables [Sh86].

In the three algorithms just mentioned the first phase is to find a good discrete structure, namely, a clique tree, a cutset, or a sequence of transformations. The goodness of the structure depends on a chosen parameter that, if selected appropriately, reduces the probabilistic computations done in the second phase. Finding a structure that optimizes the selected parameter is usually NP-hard and thus heuristic methods are applied to find a reasonable structure. Most methods in the past had no guarantee of performance and performed very badly when presented with an appropriate example. For example, the greedy algorithms of [St90] and [SC90] for the method of conditioning may in the worst case perform as bad as a factor of $n/4$ where $n$ is the number of variables in a Bayesian network. That is to say, the size of their solution instead of being 2 variables may include as many as $n/2$ variables—a disastrous outcome. Similar situations occur with other inference algorithms.

However, recently, among other results, Bar-Yehuda et al. (1994) have developed an algorithm that finds a loop cutset that is guaranteed in the worst case to contain less than 4 times the number of variables contained by a minimum loop cutset. This guarantee is given only when the number of values of every variable in the network is the same. Note that this result means that the number of instances associated with a loop cutset $F$ found by their algorithm (e.g., $r^{|F|}$ if the number of values of every variable is $r$) is no more than the number of instances associated with a minimum loop cutset raised to the forth power. Note also that, the problem of finding a minimum loop cutset was shown to be NP-hard in [SC90].

Our paper offers a new algorithm for finding a loop cutset, called MGA, that finds a loop cutset which is guaranteed in the worst case to contain less than twice the number of variables contained in an optimal loop cutset. That is, the number of instances associated with a loop cutset found by our algorithm is no more than the number of instances associated with an optimal loop cutset raised to the second power. The complexity of MGA is $O(m + n \log n)$ where $m$ and $n$ are the number of edges and vertices respectively. Unlike [BGNR94], our result holds even when the arities of the variables are arbitrary. Like [BGNR94], our solution is based on a reduction to the Weighted Vertex Feedback Set Problem, defined in the next section. We should emphasize that all these performance guarantees are for the worst case.

In Section 4 we test MGA on randomly generated graphs and find that the average ratio between the number of instances associated with the algorithms' output and the number of instances associated with a minimum solution is 1.22.

From a theoretical point of view, Bar-Yehuda et. al. (1994) note that as the number of variables grows to infinity the worst case ratio between the size of a loop cutset found by any polynomial algorithm and the size of an optimal loop cutset cannot be less than two unless the unlikely event that a similar result is obtained for the *weighted vertex cover problem (WVC)*[1]. Consequently, we conjecture that no polynomial algorithm for the loop cutset problem performs better in the worst case than the algorithm presented in this paper as graphs grow to infinity in size.

The rest of the paper is organized as follows. In Section 2 we outline the method of conditioning, explain the related loop cutset problem and describe the reduction from the loop cutset problem to the Weighted Vertex Feedback Set (WVFS) Problem. In Section 3 we provide two approximation algorithms for the WVFS problem which is by itself an NP-Complete problem [GJ79, pp. 191–192]. Finally, in Section 4 we present experiments that test the average performance of our algorithms.

## 2 The Loop Cutset Problem

Pearl's method of conditioning is one of the known inference methods for Bayesian networks. A short overview of the method of conditioning and definitions of Bayesian networks are needed. The reader is referred to [Pe88] for more details.

Let $P(u_1, \ldots, u_n)$ be a probability distribution where each $u_i$ draws values from a finite set called the *domain* of $u_i$. A directed graph $D$ with no directed cycles is called a *Bayesian network of $P$* if there is a 1–1 mapping between $\{u_1, \ldots, u_n\}$ and vertices in $D$, such that $u_i$ is associated with vertex $i$ and $P$ can be written as follows:

$$P(u_1, \ldots, u_n) = \prod_{i=1}^{n} P(u_i \mid u_{i_1}, \ldots, u_{i_{j(i)}}) \quad (1)$$

where $i_1, \ldots, i_{j(i)}$ are the source vertices of the incoming edges to vertex $i$ in $D$.

Suppose now that some variables $\{v_1, \ldots, v_l\}$ among $\{u_1, \ldots, u_n\}$ are assigned specific values $\{\mathbf{v}_1, \ldots, \mathbf{v}_l\}$ respectively. The *updating problem* is to compute the probability $P(u_i \mid v_1 = \mathbf{v}_1, \ldots, v_l = \mathbf{v}_l)$ for $i = 1, \ldots, n$.

A *trail* in a Bayesian network is a subgraph whose underlying graph is a simple path. A vertex $b$ is called a *sink* with respect to a trail $t$ if there exist two consecutive edges $a \rightarrow b$ and $b \leftarrow c$ on $t$. A trail $t$ is *active by a set of vertices $Z$* if (1) every sink with respect to $t$ either is in $Z$ or has a descendant in $Z$ and (2) every other vertex along $t$ is outside $Z$. Otherwise, the trail is said to be *blocked (d-separated)* by $Z$.

Verma and Pearl [VP88] have proved that if $D$ is a Bayesian network of $P(u_1, \ldots, u_n)$ and all trails between a vertex in $\{r_1, \ldots, r_l\}$ and a vertex in $\{s_1, \ldots, s_k\}$ are blocked by $\{t_1, \ldots, t_m\}$, then the corresponding sets of variables $\{u_{r_1}, \ldots, u_{r_l}\}$ and $\{u_{s_1}, \ldots, u_{s_k}\}$ are independent conditioned on $\{u_{t_1}, \ldots, u_{t_m}\}$. Furthermore, Geiger and Pearl [GP90] proved a converse to this theorem. Both results are presented and extended in [GVP90].

Using the close relationship between blocked trails and conditional independence, Kim and Pearl [KP83] developed an algorithm UPDATE-TREE that solves the updating problem on Bayesian networks in which every two vertices are connected with at most one trail (singly-connected). Pearl then solved the updating problem on any Bayesian network as follows [Pe86]. First, a set of vertices $S$ is selected such that any two vertices in the network are connected by at most one *active* trail in $S \cup Z$, where $Z$ is any subset of vertices. Then, UPDATE-TREE is applied once for each combination of value assignments to the variables corresponding to $S$, and, finally, the results are combined. This algorithm is called the method of *conditioning* and its complexity grows exponentially with the size of $S$. The set $S$ is called a *loop cutset*. Note that when the domain size of the variables varies, then UPDATE-TREE is called a number of times equal to the product of the domain sizes of the variables whose corresponding vertices participate in the loop cutset. If we take the logarithm of the domain size (number of values) as the weight of a vertex, then finding a loop cutset such that the sum of its vertices weights is minimum optimizes Pearl's updating algorithm in the case where the domain sizes may vary.

We now give an alternative definition for a loop cutset $S$ and then provide an approximation algorithm for finding it. This definition is borrowed from [BGNR94]. The underlying graph $G$ of a directed graph $D$ is the undirected graph formed by ignoring the directions of the edges in $D$. A *cycle* in $G$ is a path whose two terminal vertices coincide. A *loop* in $D$ is a subgraph of $D$ whose underlying graph is a cycle. A vertex $v$ is a *sink* with respect to a loop $\Gamma$ if the two edges adjacent to $v$ in $\Gamma$ are directed into $v$. Every loop must contain at least one vertex that is not a sink with respect to that loop. Each vertex that is not a sink with respect to a loop $\Gamma$ is called an *allowed vertex with respect to $\Gamma$*. A *loop cutset* of a directed graph $D$ is a set of vertices that contains at least one allowed vertex with respect to each loop in $D$. The weight of a set of vertices $X$ is denoted by $w(X)$ and is equal to $\sum_{v \in X} w(v)$ where $w(x) = \log(|x|)$ and $|x|$ is the size of the domain associated with vertex $x$. A *minimum loop cutset* of a weighted directed graph $D$ is a loop cutset $F^*$ of $D$ for which $w(F^*)$ is minimum over all loop cutsets of $G$. The *Loop Cutset Problem* is defined as finding a minimum loop cutset of a given weighted directed graph $D$.

The approach we take is to reduce the weighted loop cutset problem to the weighted vertex feedback set problem, as done by [BGNR94]. We now define the weighted vertex feedback set problem and then the reduction.

---

[1] The WVC problem is finding a set of vertices that contains an endpoint of every edge in a given undirected graph and which has a minimum weight among all such sets.

Let $G = (V, E)$ be an undirected graph, and let $w : V \to {\rm I\!R}^+$ be a weight function on the vertices of $G$. A *vertex feedback set* of $G$ is a subset of vertices $F \subseteq V$ such that each cycle in $G$ passes through at least one vertex in $F$. In other words, a vertex feedback set $F$ is a set of vertices of $G$ such that by removing $F$ from $G$, along with all the edges incident with $F$, we obtain a set of trees (i.e., a forest). The weight of a set of vertices $X$ is denoted (as before) by $w(X)$ and is equal to $\sum_{v \in X} w(v)$. A *minimum vertex feedback set* of a weighted graph $G$ with a weight function $w$ is a vertex feedback set $F^*$ of $G$ for which $w(F^*)$ is minimum over all vertex feedback sets of $G$. The *Weighted Vertex Feedback Set (WVFS) Problem* is defined as finding a minimum vertex feedback set of a given weighted graph $G$ having a weight function $w$. Application of this problem for constraint satisfaction is described in [DP90].

In the next section we offer an algorithm, called MGA, for approximately solving the weighted vertex feedback set problem. The algorithm is guaranteed to output a weighted vertex set whose weight is less than twice the optimal weight.

The reduction is as follows. Given a weighted directed graph $(D, w)$ (e.g., a Bayesian network), we define the *splitting* weighted undirected graph $D_s$ with a weight function $w_s$ as follows. Split each vertex $v$ in $D$ into two vertices $v_{\rm in}$ and $v_{\rm out}$ in $D_s$ such that all incoming edges to $v$ in $D$ become undirected incident edges with $v_{\rm in}$ in $D_s$, and all outgoing edges from $v$ in $D$ become undirected incident edges with $v_{\rm out}$ in $D_s$. In addition, connect $v_{\rm in}$ and $v_{\rm out}$ in $D_s$ by an undirected edge. Now set $w_s(v_{\rm in}) = \infty$ and $w_s(v_{\rm out}) = w(v)$. For a set of vertices $X$ in $D_s$, we define $\psi(X)$ as the set obtained by replacing each vertex $v_{\rm in}$ or $v_{\rm out}$ in $X$ by the respective vertex $v$ in $D$ from which these vertices originated.

Our algorithm can now be easily stated.

**Algorithm LC**
*Input:* A Bayesian network $D$;
*Output:* A loop cutset of $D$;

1. Construct the splitting graph $D_s$ with weight function $w_s$;

2. Apply MGA on $(D_s, w_s)$ to obtain a vertex feedback set $F$;

3. Output $\psi(F)$.

It is immediately seen that if MGA outputs a vertex feedback set $F$ whose weight is no more than twice the weight of a minimum vertex feedback set of $D_s$, then $\psi(F)$ is a loop cutset of $D$ with weight no more than twice the weight of a minimum loop cutset of $D$. This observation holds because there is an obvious one-to-one and onto correspondence between loops in $D$ and cycles in $D_s$ and because MGA never chooses a vertex that has an infinite weight.

## 3 Algorithms For The WVFS problem

Recall that the weighted vertex feedback set problem is defined as finding a minimum vertex feedback set of a given weighted graph $G$.

### 3.1 The Greedy Algorithm

We first analyze the simplest of all approximation algorithms for the weighted vertex feedback set problem—the greedy algorithm. Assume we are given a weighted undirected graph $G$ with a weight function $w$. The greedy algorithm starts with $G$ after removing all vertices with degree 0 or 1 and repeatedly chooses to insert a vertex $v$ into the constructed vertex feedback set if the ratio between $v$'s weight $w(v)$ and $v$'s degree $d(v)$ in the current graph is minimal across all vertices in the current graph. When $v$ is selected, it is removed from the current graph and then all vertices with degree 0 or 1 are repeatedly removed as well. This step is repeated until the graph is exhausted.

This algorithm and parts of its analysis are influenced by the work of Chvatal (1979) who analyzed the greedy algorithm for the Weighted Set Cover problem (WSC) and by Lovász (1975) and Johnson (1974) who analyzed the unweighted version of this problem.

**ALGORITHM GA**

> **Input:** *A weighted undirected graph $G(V, E, w)$.*
> **Output:** *A vertex feedback set $F$.*
> $\quad F \leftarrow \emptyset$
> $\quad i \leftarrow 1$
> Repeatedly remove all vertices with
> $\quad$ degree 0 or 1 from $V$ and insert
> $\quad$ the resulting graph into $G_i$
> **While** $G_i$ is not the empty graph **do**
> $\quad$ 1. Pick a vertex $v_i$ for which
> $\quad\quad \frac{w(v_i)}{d(v_i)}$ is minimum in $G_i$
> $\quad$ 2. $F \leftarrow F \cup \{v_i\}$
> $\quad$ 3. $V \leftarrow V \setminus \{v_i\}$
> $\quad$ 4. $i \leftarrow i + 1$
> $\quad$ 5. Repeatedly remove all vertices
> $\quad\quad$ with degree 0 or 1 from $V$
> $\quad\quad$ and insert the resulting
> $\quad\quad$ graph into $G_i$
> **end.**

In the rest of this section we prove that the performance ratio of this greedy algorithm is bounded by $2(\log d + 1)$ where $d = max_{v \in V} d(v)$ is the degree of the graph. Recall that the performance ratio of an approximation algorithm is the worst case ratio between the weight of the algorithm's output and the weight of an optimal solution. In Section 4, we show experimentally that even this simple algorithm when combined with the reduction algorithm LC convincingly outperforms the algorithms given by [SC90, St90].

Let $F^*$ be an optimal weighted feedback set of $G(V, E, w)$ and let $\overline{F^*} = V \setminus F^*$. Note that the vertices in $F$ (the output of GA) are denoted by $\{v_1, v_2, \ldots, v_t\}$ where $v_i$ are indexed in the order in which they are inserted into $F$ by GA and where $t = |F|$. Let $d_i(v)$ denote the degree of vertex $v$ in $G_i$—the graph generated in iteration $i$ of GA—and let $V_i$ be the set of vertices of $G_i$. An edge is *covered* by the algorithm if for some $i = 1, \ldots, t$, one of its endpoints is $v_i$ and the edge exists in $G_i$. Let $\Gamma_1(v)$ denote the set of edges in $G_1$ for which at least one endpoint is $v$. Note that the set of vertex feedback sets of $G$ and $G_1$ is the same and that the degree of every vertex in $G_1$ is smaller or equal to the degree of that vertex in $G$.

Let $c_i = w(v_i)/d_i(v_i)$ and let $C(e) = c_i$ for every edge $e$ removed at iteration $i$. Note that for every $j \leq i$ we have $w(v_j)/d_j(v_j) \leq w(v_i)/d_i(v_i)$ because vertices are selected in decreasing order of these ratios. Also note that for $j \leq i$, $d_j(v_i) \geq d_i(v_i)$ since the algorithm never adds edges. Thus,

$$c_j \equiv w(v_j)/d_j(v_j) \leq w(v_i)/d_i(v_i) \equiv c_i \quad (2)$$

for $1 \leq j \leq i \leq |F|$, as originally claimed by [Ch79] in the context of the WSC problem.

To analyze the performance ratio we use a lemma that bounds the number of edges in $G_i$ covered by the algorithm until its termination. We need the following definitions. Let $d_X(v)$ be the number of edges whose one endpoint is $v$ and the other is a vertex in $X$. Denote $F_i^* = F^* \cap V_i$ and $\overline{F_i^*} = \overline{F^*} \cap V_i$. A *linkpoint* is a vertex that has a degree 2 and A *branchpoint* is a vertex that has a degree larger than 2. (A self-loop adds 2 to the degree of a vertex).

**Lemma 1**

$$\sum_{j=i}^{t} d_j(v_j) \leq 2 \sum_{v \in F_i^*} d_i(v), \quad (3)$$

**Proof:** We will actually prove that,

$$\sum_{j=i}^{t} d_j(v_j) \leq \sum_{v \in V_i} (d_i(v) - 2) + 2|F_i^*| \leq 2 \sum_{v \in F_i^*} d_i(v). \quad (4)$$

According to our notations, $\sum_{v \in V_i}(d_i(v) - 2)$ equals

$$\sum_{v \in \overline{F_i^*}} (d_{\overline{F_i^*}}(v) - 2) + \sum_{v \in \overline{F_i^*}} d_{F_i^*}(v) + \sum_{v \in F_i^*} (d_i(v) - 2).$$

Furthermore, the graph induced by $\overline{F_i^*}$ is a forest and since the number of edges in a forest is smaller (or equal) than the number of vertices, we have, $\sum_{v \in \overline{F_i^*}} d_{\overline{F_i^*}}(v)/2 \leq |\overline{F_i^*}|$. Thus $\sum_{v \in \overline{F_i^*}}(d_{\overline{F_i^*}}(v) - 2) \leq 0$. Consequently, $\sum_{v \in V_i}(d_i(v) - 2) + 2|F_i^*|$ is less than or equal to

$$\sum_{v \in \overline{F_i^*}} d_{F_i^*}(v) + \sum_{v \in F_i^*} d_i(v) \leq 2 \sum_{v \in F_i^*} d_i(v).$$

The proof of the first part of Eq. 4 is constructive. We repeatedly apply the following procedure on $G_i$ selecting in each step a vertex $v_j \in F_i$ and showing that there are terms in the right hand side (RHS) of Eq. 4 that contribute $d_j(v_j)$ to the RHS and have not been used for any other $v \in F_i$. Set $H = G_i$ and for $k = i \ldots t$ do as follows:

Pick the vertex $v_k$. If $v_k$ is a linkpoint in $H$ then follow the two paths $p_1$ and $p_2$ in $H$ emanating from $v_k$ until the first branchpoint on each side is found. There are three cases to consider. Either two distinct branchpoints $b_1$ and $b_2$ are found, one branchpoint $b_1$ (in which case $p_1$ and $p_2$ define a cycle) or none (if the cycle is isolated). In the first case the two edges on $p_1$ and $p_2$ whose endpoints are $b_1$ and $b_2$, respectively, are associated with the terms $d_k(b_1) - 2 > 0$ and $d_k(b_2) - 2 > 0$ in the RHS and so each of these terms contributes 1 to the sum $\sum_{v \in V_i}(d_i(v) - 2)$. In the second case, similarly, the two edges on $p_1$ and $p_2$ whose endpoints is $b_1$ are associated with the term $d_k(b_1) - 2 > 0$ and so, if $d_k(b_1) > 3$, this term contributes 2 to the sum $\sum_{v \in V_i}(d_i(v) - 2)$. If $d_k(b_1) = 3$ we continue to follow the third path from $b_1$ (i.e., not $p_1$ or $p_2$) until another branchpoint $b_2$ is found and the last edge on that path is associated with $d_k(b_2) - 2$ which contributes the extra missing 1 to the RHS. Finally, if no branchpoint is found, then on the cycle in which $v_k$ resides there must exist a vertex from $F_i^*$ that resides on no other cycles of $H$. Now, if $v_k$ is a branchpoint, then the term $d_k(v_k) - 2$ appears in both sides of the inequality. In this case, sequentially remove $d_k(v_k) - 2$ of the $d_k(v_k)$ edges adjacent to $v_k$ such that after each removal the vertices with degree 0 or 1 are removed from $H$ as well. Thus, $v_k$ remains a linkpoint in which case the procedure for a linkpoint is applied. Finally, remove $v_k$, and repeatedly remove all the vertices with degree 0 or 1 from $H$. Repeat until $F_i$ is exhausted. □

We now show that $w(F) \leq 2 \cdot (\log d + 1) \cdot w(F^*)$.

$$w(F) = \sum_{i=1}^{t} w(v_i) = \sum_{i=1}^{t} c_i \cdot d_i(v_i) =$$

$$c_1 \sum_{i=1}^{t} d_i(v_i) + \sum_{i=2}^{t} (c_i - c_{i-1}) \sum_{j=i}^{t} d_j(v_j) \quad (5)$$

Since $c_i \geq c_{i-1}$, we can apply Eq. 3 and so,

$$w(F) \leq 2c_1 \sum_{v \in F_1^*} d_1(v) + \sum_{i=2}^{t} 2(c_i - c_{i-1}) \sum_{v \in F_i^*} d_i(v) =$$

$$\sum_{i=1}^{t} 2c_i \sum_{v \in F_i^*} d_i(v) - \sum_{i=1}^{t-1} 2c_i \sum_{v \in F_{i+1}^*} d_{i+1}(v)$$

Thus,

$$w(F) \leq \sum_{i=1}^{t} 2c_i \sum_{v \in F_i^* \setminus F_{i+1}^*} d_i(v) +$$

$$\sum_{i=1}^{t} 2c_i \sum_{v \in F_{i+1}^*} d_i(v) - \sum_{i=1}^{t-1} 2c_i \sum_{v \in F_{i+1}^*} d_{i+1}(v) =$$

$$2(\sum_{i=1}^{t-1} \left( c_i \sum_{v \in F_i^* \setminus F_{i+1}^*} d_i(v) + c_i \sum_{v \in F_{i+1}^*} (d_i(v) - d_{i+1}(v)) \right)$$

$$+ c_t \sum_{v \in F_t^*} d_t(v))$$

However, since the last sum on the right hand side merely counts the edge weights according to the iteration they are assigned a weight, we get,

$$w(F) \leq 2 \sum_{v \in F^*} \sum_{e \in \Gamma_1(v)} C(e) \qquad (6)$$

Now, for every $v \in F^*$,

$$H(d(v)) \cdot w(v) \geq \sum_{e \in \Gamma_1(v)} C(e), \qquad (7)$$

where $H(m) = \sum_{i=1}^{m} 1/i$, as shown in [Ch79] using the following argument. Let $s$ be the largest superscript such that $d_s(v) > 0$ then

$$\sum_{e \in \Gamma_1(v)} C(e) = \sum_{i=1}^{s} (d_i(v) - d_{i+1}(v)) \cdot (w(v_i)/d_i(v_i))$$

$$\leq w(v) \sum_{i=1}^{s} (d_i(v) - d_{i+1}(v))/d_i(v)$$

where the inequality is due to Eq. 2. Furthermore, by induction,

$$\sum_{e \in \Gamma_1(v)} C(e) \leq w(v) \sum_{i=1}^{s} [H(d_i(v)) - H(d_{i+1}(v))].$$

Since the right hand side is equal to $w(v) \cdot H(d(v))$, Eq. 7 follows. Combining Eqs. 6, and 7 yields,

$$w(F) \leq 2 \sum_{v \in F^*} H(d(v)) \cdot w(v) \leq 2H(d) \cdot w(F^*).$$

Thus, since $H(d) \leq \log d + 1$ (equality holds only when $d = 1$),

**Theorem 2** *The performance ratio of* GA *is bounded by* $2(\log d + 1)$.

We have an example in which the ratio between GA's output and the optimal output is $2 \log d$. Our example is similar to the example for the vertex cover problem given in [Mo92, pp. 47]. Consequently, the upper bound given in Theorem 2 is rather tight.

### 3.2 The Modified Greedy Algorithm

We now present a modified greedy algorithm, called MGA, whose performance ratio is bounded by the constant 2. The changes we introduce into the greedy algorithm are quite minor and so it is interesting that such a vast improvement in the performance ratio is obtained. A similar phenomenon is reported in the context of the weighted vertex cover problem [Cl83].

MGA has two phases. In the first phase MGA repeatedly chooses to insert a vertex $v$ into the constructed vertex feedback set if the ratio between $v$'s weight $w(v)$ and $v$'s degree $d(v)$ in the current graph is minimal across all vertices in the current graph. When $v$ is selected, it is removed from the current graph and then all vertices with degree 0 or 1 are repeatedly removed as well. For every edge removed in this process, a weight of $w(v)/d(v)$ is subtracted from its endpoint vertices. These steps are repeated until the graph is exhausted. The only difference between this phase and the plain greedy algorithm is the revision of some weights in each step instead of just revising the current degrees. The second phase removes redundant vertices from the constructed vertex feedback set.

**ALGORITHM MGA**

```
Input:  A weighted undirected graph G(V, E, w).
Output: A vertex feedback set F.
    F' ← ∅
    i ← 1
    Repeatedly remove all vertices with degree 0
        or 1 from V and their adjacent edges from
        E and insert the resulting graph into G_i.
    While G_i is not the empty graph do
        1. Pick a vertex v_i for which
               w(v_i)/d(v_i) is minimum in G_i
        2. F' ← F' ∪ {v_i}
        3. V ← V \ {v_i}
        4. i ← i + 1
        5. Repeatedly remove all vertices with
               degree 0 or 1 from V and their
               adjacent edges from E and insert
               the resulting graph into G_i.
           For every edge e = (u_1, u_2)
               removed in this process do
                   C(e) ← w(v_i)/d(v_i)
                   w(u_1) ← w(u_1) - C(e)
                   w(u_2) ← w(u_2) - C(e)
    end
    F ← F'
    For i = |F| to 1 do {Phase 2}
        If every cycle in G_i that intersects
               with {v_i} also intersects
               with F \ {v_i} then,
            F ← F \ {v_i}
    endfor
end
```

Clearly $F'$ computed at the first phase of MGA is a vertex feedback set of $G$ and $F$ created from $F'$ by removing all redundant vertices is a *minimal vertex feedback set* of $G$, that is, if a vertex is removed from $F$, then $F$ ceases to be a vertex feedback set of $G$. Furthermore, as a result of removing redundant vertices the inequality $\sum_{j=i}^{t} d_j(v_j) \leq 2\sum_{v \in F_i^*} d_i(v)$ (Eq. 3), proven to hold for the greedy algorithm becomes,

$$\sum_{v \in F_i} d_i(v) \leq 2 \sum_{v \in F_i^*} d_i(v), \tag{8}$$

where $F_i^*$ are the vertices in $F$ that appear in graph $G_i$. The proof of this equation is postponed to Section 3.3. From the description of the algorithm we have for every vertex $v$ in $G_1$,

$$\sum_{e \in \Gamma_1(v)} C(e) \leq w(v) \tag{9}$$

and if $v \in F$ equality must hold. Eq. 9 replaces the inequality $\sum_{e \in \Gamma_1(v)} C(e) \leq H(d(v)) \cdot w(v)$ (Eq. 7) proven for the greedy algorithm. By analogy with the previous section and using similar lines of reasoning, it is clear that Eqs. 8 and 9 which replace Eqs. 3 and 7 show that the bound on the performance ratio drops from $2 \cdot H(d)$ for the greedy algorithm to 2 for the modified greedy algorithm.

**Theorem 3** *Algorithm* MGA *always outputs a vertex feedback set whose weight is no more than twice the weight of the optimal vertex feedback set.*

**Proof.** As in Section 3.1, $F^*$ denotes a minimum feedback set of $G(V, E, w)$ and $\overline{F^*} = V \setminus F^*$. Recall that the vertices in the constructed set $F'$ are $\{v_1, v_2, \ldots, v_t\}$ where $v_i$ are indexed in the order in which they are inserted into $F$ by MGA and $t = |F'|$. Also, $w_i(v)$ and $d_i(v)$ denote the weight and degree, respectively, of vertex $v$ in $G_i$—the graph generated in iteration $i$ of Step 5 of MGA—and $V_i$ denotes the set of vertices of $G_i$.

As in the greedy algorithm, for every $j \leq i$ we have $w_j(v_j)/d_j(v_j) \leq w_j(v_i)/d_j(v_i)$ and also $w_j(v_i)/d_j(v_i) \leq w_i(v_i)/d_i(v_i)$ due to the way that the current weights and degrees are updated in the algorithm. Thus,

$$c_j \equiv w_j(v_j)/d_j(v_j) \leq w_i(v_i)/d_i(v_i) \equiv c_i \tag{10}$$

for $1 \leq j \leq i \leq |F'|$.

We also have,

$$\sum_{e \in \Gamma_1(v_i)} C(e) = c_i \cdot d_i(v_i) + \sum_{j=1}^{i-1} c_j \cdot (d_j(v_i) - d_{j+1}(v_i)) \tag{11}$$

because the right hand side simply groups edges according to the iteration in which they are assigned a weight.

Let $\alpha_i = 1$ if $v_i \in F$ and $\alpha_i = 0$ if $v_i \notin F$. That is, $\alpha_i$ is 1 if $v_i$ is not removed from $F$ in the final stage of MGA and 0 otherwise. We now prove that $w(F) \leq 2 \cdot w(F^*)$.

$$w(F) = \sum_{i=1}^{t} \alpha_i \cdot w(v_i) = \sum_{i=1}^{t} \alpha_i \sum_{e \in \Gamma_1(v_i)} C(e)$$

Now, due to Eq. 11, $w(F)$ is equal to

$$\sum_{i=1}^{t} \alpha_i \cdot \left[ c_i \cdot d_i(v_i) + \sum_{j=1}^{i-1} c_j \cdot (d_j(v_i) - d_{j+1}(v_i)) \right]$$

which in turn equals to

$$c_1 \sum_{i=1}^{t} \alpha_i \cdot d_1(v_i) + \sum_{i=2}^{t}(c_i - c_{i-1}) \sum_{j=i}^{t} \alpha_j \cdot d_i(v_j)$$

Furthermore,

$$\sum_{j=i}^{t} \alpha_j \cdot d_i(v_j) = \sum_{v \in F_i} d_i(v) \leq 2 \sum_{v \in F_i^*} d_i(v). \tag{12}$$

Since $c_i \geq c_{i-1}$, we can apply Eq. 12 and so, analogously to the derivation of Eq. 6, we get,

$$w(F) \leq$$

$$2c_1 \sum_{v \in F_1^*} d_1(v) + \sum_{i=2}^{t} 2(c_i - c_{i-1}) \sum_{v \in F_i^*} d_i(v) \leq$$

$$2 \sum_{v \in F^*} \sum_{e \in \Gamma_1(v)} C(e) \tag{13}$$

Now, Eqs. 9 and 13 yield the claimed inequality, $w(F) \leq 2 \sum_{v \in F^*} w(v) = 2w(F^*)$. □

The complexity of the first phase of MGA is $O(|E| + |V| \log |V|)$ using a Fibonacci heap (e.g., [FT87]) because finding and deleting a vertex with minimum ratio $w(v)/d(v)$ from the heap is done $|V|$ times at the cost of $O(\log |V|)$ and decreasing a weight from a vertex in the heap is done $|E|$ times at an amortized cost of $O(1)$. The complexity of the second phase of MGA is also is $O(|E| + |V| \log |V|)$ using a simple implementation of the union-find algorithm because we need to do at most $|V|$ union operations at an amortized cost of $O(\log |V|)$ and at most $|E|$ find operations at the cost of $O(1)$ [CLR90, pp. 445].

Interestingly, if the second phase is removed from MGA (making MGA even closer to GA), then it can be shown that the performance ratio becomes 4 rather than 2. Hence the vast improvement in the worst-case performance of MGA compared to GA stems from changing the vertices' weights in each step rather than from removing redundant vertices.

### 3.3 A Theorem about Minimal Vertex Feedback Sets

In this section we prove Eq. 8 which has been used in the analysis of the modified greedy algorithm. Let

$G$ be a weighted graph for which every vertex has a degree strictly greater than 1, $F$ be a minimal vertex feedback set of $G$ and $F^*$ be an arbitrary vertex feedback set of $G$ (possibly a minimum weight vertex feedback set). Let $d(v)$ be the degree of vertex $v$ and $d_X(v)$ be the number of edges whose one endpoint is $v$ and the other is in a set of vertices $X$.

**Theorem 4** *Let $G, F$ and $F^*$ be defined as above. Then, $\sum_{v \in F} d(v) \leq 2 \sum_{v \in F^*} d(v)$.*

This theorem is interesting by its own sake since it relates the number of edges adjacent to any minimal weighted vertex feedback set to the number of edges adjacent to any minimum weighted vertex feedback set. Note that $F_i^*$ is a minimal vertex feedback set of $G_i$ and therefore Theorem 4 proves Eq. 8.

To prove this theorem we divide $\sum_{v \in F} d(v)$ into the sum $2|F| + \sum_{v \in F}(d(v)-2)$ and provide an upper bound for each term.

**Lemma 5** *Let $G, F$ and $F^*$ be defined as above. Then,*

$$2|F| \leq \sum_{v \in \overline{F}} d(v) - 2|\overline{F} \cap \overline{F^*}| + 2|F \cap F^*| \quad (14)$$

**Proof:** First note that for every set of vertices $B$ in $G$,

$$\sum_{v \in \overline{F}} d(v) - 2|\overline{F} \cap \overline{F^*}| = \sum_{v \in \overline{F} \cap B} d(v) + \sum_{v \in \overline{F} \setminus B} d(v) - 2|\overline{F} \cap \overline{F^*} \cap B| - 2|(\overline{F} \cap \overline{F^*}) \setminus B| \quad (15)$$

However, the degree of every vertex in $G$ satisfies $d(v) \geq 2$ and therefore $\sum_{v \in \overline{F} \setminus B} d(v) \geq 2|(\overline{F} \cap \overline{F^*}) \setminus B|$. Consequently,

$$\sum_{v \in \overline{F}} d(v) - 2|\overline{F} \cap \overline{F^*}| \geq \sum_{v \in \overline{F} \cap B} d(v) - 2|\overline{F} \cap \overline{F^*} \cap B|. \quad (16)$$

Thus, and since $|F \cap F^*| \geq |F \cap F^* \cap B|$ and $d_B(v) \leq d(v)$, to prove the lemma it suffices to show that

$$2|F| \leq \sum_{v \in \overline{F} \cap B} d_B(v) - 2|\overline{F} \cap \overline{F^*} \cap B| + 2|F \cap F^* \cap B|, \quad (17)$$

or equivalently,

$$2|F| \leq \sum_{v \in \overline{F} \cap B} (d_B(v) - 2) + 2|F^* \cap B|, \quad (18)$$

holds for some set of vertices $B$. We now define a set $B$ for which this inequality can be proven. Since $F$ is minimal, each vertex in $F$ can be associated with a cycle in $G$ that contains no other vertices of $F$. We define a graph $H$ that consists of the union of these cycles—one cycle per each vertex. Note that every vertex in $F$ is a linkpoint in $H$, i.e., a vertex with degree 2. Let $B$ be the vertices of $H$.

The proof of Eq. 18 is constructive. We repeatedly apply the following procedure on $H$ selecting in each step a vertex $v \in F$ and showing that there are terms in the right hand side (RHS) of Eq. 18 that contribute 2 to the RHS and have not been used for any other $v \in F$.

Set $H' = H$. Pick a vertex $v \in F$ and follow the two paths $p_1$ and $p_2$ in $H'$ emanating from $v$ (which is a linkpoint) until the first branchpoint on each side is found. There are three cases to consider. Either two distinct branchpoints $b_1$ and $b_2$ are found, one branchpoint $b_1$ (in which case $p_1$ and $p_2$ define a cycle) or none (if the cycle is isolated). In the first case the two edges on $p_1$ and $p_2$ whose endpoints are $b_1 \in \overline{F}$ and $b_2 \in \overline{F}$, respectively, are associated with the terms $d_B(b_1) - 2 > 0$ and $d_B(b_2) - 2 > 0$ in the RHS and so each of these terms contributes 1 to the sum $\sum_{v \in \overline{F} \cap B}(d_B(v) - 2)$. In the second case, similarly, the two edges on $p_1$ and $p_2$ whose endpoints is $b_1 \in \overline{F}$ are associated with the term $d_B(b_1) - 2 > 0$ and so, if $d_B(b_1) > 3$, this term contributes 2 to the sum $\sum_{v \in \overline{F} \cap B}(d_B(v) - 2)$. If $d_B(b_1) = 3$ we continue to follow the third path from $b_1$ (i.e., not $p_1$ or $p_2$) until another branchpoint $b_2 \in \overline{F}$ is found and the last edge on that path is associated with $d_B(b_2) - 2$ which contributes the extra missing 1 to the RHS. Finally, if no branchpoint is found, then on the cycle in which $v$ resides there must exist a vertex from $F^*$ that resides on no other cycles of $H'$. Thus, the third case could not occur more than $|F^* \cap B|$ times. Now remove the paths $p_1$ and $p_2$ from $H'$ obtaining a graph in which still each vertex in $F$ resides on a cycle that contains no other vertices of $F$. Continue the process until $F$ is exhausted. □

**Lemma 6** *Let $G, F$ and $F^*$ be defined as above. Then the sum $\sum_{v \in F}(d(v) - 2)$ is upper bounded by,*

$$\sum_{v \in F \cap \overline{F^*}} d_{F^*}(v) + \sum_{v \in F \cap F^*}(d(v) - 2) - \sum_{v \in \overline{F} \cap \overline{F^*}}(d_{\overline{F^*}}(v) - 2)$$

**Proof:** First note that,

$$\sum_{v \in F}(d(v) - 2) = \sum_{v \in F \cap \overline{F^*}}(d_{\overline{F^*}}(v) - 2) + \sum_{v \in F \cap \overline{F^*}} d_{F^*}(v) + \sum_{v \in F \cap F^*}(d(v) - 2) + \sum_{v \in \overline{F} \cap \overline{F^*}}(d_{\overline{F^*}}(v) - 2) - \sum_{v \in \overline{F} \cap \overline{F^*}}(d_{\overline{F^*}}(v) - 2) \quad (19)$$

We now claim that $\sum_{v \in F \cap \overline{F^*}}(d_{\overline{F^*}}(v) - 2) + \sum_{v \in \overline{F} \cap \overline{F^*}}(d_{\overline{F^*}}(v) - 2)$ is less or equal than 0 and therefore can be omitted from the inequality and conclude this proof. The graph induced by $\overline{F^*}$ is a forest and since the number of edges in a forest is smaller than the number of vertices, we have, $\sum_{v \in \overline{F^*}} d_{\overline{F^*}}(v)/2 \leq |\overline{F^*}|$. Thus $\sum_{v \in \overline{F^*}}(d_{\overline{F^*}}(v) - 2) \leq 0$ which is equivalent to the stated claim. □

Using the bounds given by Lemmas 5 and 6 we have,

$$\sum_{v \in F} d(v) \leq \sum_{v \in \overline{F}} d(v) - 2|\overline{F} \cap \overrightarrow{F^*}| +$$

$$2|F \cap F^*| + \sum_{v \in F \cap \overline{F^*}} d_{F^*}(v)$$

$$+ \sum_{v \in F \cap F^*} (d(v) - 2) - \sum_{v \in \overline{F} \cap \overline{F^*}} (d_{\overline{F^*}}(v) - 2)$$

However, $\sum_{v \in F \cap F^*}(d(v) - 2) + 2|F \cap F^*| = \sum_{v \in F \cap F^*} d(v)$ and $\sum_{v \in \overline{F} \cap \overline{F^*}}(d_{\overline{F^*}}(v) - 2) + 2|\overline{F} \cap \overrightarrow{F^*}| = \sum_{v \in \overline{F} \cap \overline{F^*}} d_{\overline{F^*}}(v)$. Thus, $\sum_{v \in F} d(v)$ is bounded by

$$\sum_{v \in \overline{F}} d(v) + \sum_{v \in F \cap \overline{F^*}} d(v) -$$

$$\sum_{v \in \overline{F} \cap \overline{F^*}} d_{\overline{F^*}}(v) + \sum_{v \in F \cap \overline{F^*}} d_{F^*}(v).$$

Now, $\sum_{v \in \overline{F}} d(v) - \sum_{v \in \overline{F} \cap \overline{F^*}} d_{\overline{F^*}}(v)$ actually equals to $\sum_{\overline{F} \cap F^*} d(v) + \sum_{v \in \overline{F} \cap \overline{F^*}} d_{F^*}(v)$ and therefore

$$\sum_{v \in F} d(v) \leq \sum_{v \in \overline{F^*}} d_{F^*}(v) + \sum_{v \in F^*} d(v) \leq 2 \sum_{v \in F^*} d(v)$$

which concludes the proof of Theorem 4.

## 4   Experimental Results

Below we denote by A1 the algorithm described in [SC90] and by A2 the algorithm described in [St90]. We performed six experiments. In the first two experiments we tested how the outputs of the four algorithms, A1, A2, GA, and MGA, compare to a minimum loop cutset. In two additional experiments we checked how the algorithms' outputs compare to each other when given larger graphs for which a minimum loop cutset is hard to obtain. In the above four experiments we have chosen all variables to be binary. The final two experiments compare the performance of these algorithms when the number of values in each vertex is randomly chosen between 2 and 6, 2 and 8, and between 2 and 10. Each instance of the six experiments is based on 100 graphs generated as described by [SC90].

In the first experiment each of the 100 graphs generated had 15 vertices and 25 edges. MGA made only one mistake producing 6 vertices instead of the minimum of 5 vertices. GA made 4 mistakes each by one vertex off. A2 made 7 mistakes one of which was two vertices off the minimum and the other six mistakes were one vertex off. A1 made 11 mistakes one of which was 2 vertices off and the other 10 mistakes were one vertex off. The minimum loop cutsets were between 3 and 6 vertices. Note that the ratio between the number of instances associated with a loop cutset found by MGA in this experiment and the number of instances associated with a minimum loop cutset is 1.002 which is far less than the theoretical ratios guaranteed by Theorem 4 for this experiment which lie between 8 when the minimum loop cutset contains 3 binary variables and 64 when the minimum loop cutset contains 6 binary variables.

In the second experiment we generated 100 networks each with 25 vertices and 25 edges and tested how the output of the four algorithms compare to a minimum loop cutset when the graphs have a small number of loops. This case is interesting because the conditioning inference algorithm is most appropriate for these networks. MGA made no mistakes while the other three algorithms made between 4 and 5 mistakes each by one vertex (the minimum loop cutsets contained between 2 and 4 vertices).

Next we tested larger graphs. The first portion of the table below compares between GA and A2 showing that GA performs better than A2 in 53 of the 61 graphs (87%) in which the algorithms disagree (out of 600 graphs tested). Each line in the table is based on 100 randomly generated graphs. The output columns show the number of graphs for which the two algorithms had an output of the same size and the number of graphs each algorithm performed better than the other. Thus even our simple greedy algorithm GA performs much better than A2. The reason for this is the reduction from the loop cutset problem to the weighted vertex feedback set problem which allows the algorithm to select vertices that have parents while A2 unjustifiably does not select such vertices (unless they have no pair of parents residing on the same loop). Similar empirical results and the same explanation applies to A1. The second portion of the table shows that MGA performs better than GA in 67 of the 75 graphs (89%) in which the algorithms disagreed. Comparing MGA and A2 in the same fashion (600 graphs) showed that MGA performed better than A2 in 109 of the 116 graphs in which the algorithms disagreed. Similarly, MGA performed better than A1 in 135 of the 137 graphs in which these algorithms disagreed.

| $|V|$ | $|E|$ | A2 | GA | Eq. | GA | MGA | Eq. |
|---|---|---|---|---|---|---|---|
| 25 | 25 | 0 | 1 | 99 | 0 | 4 | 96 |
| 25 | 50 | 1 | 8 | 91 | 0 | 8 | 92 |
| 25 | 75 | 0 | 15 | 85 | 1 | 7 | 92 |
| 55 | 55 | 1 | 2 | 97 | 0 | 9 | 91 |
| 55 | 75 | 4 | 10 | 86 | 1 | 18 | 83 |
| 55 | 105 | 2 | 17 | 81 | 6 | 21 | 83 |
|  |  | 8 | 53 | 539 | 8 | 67 | 525 |

Finally, we repeated some of the experiments except that now each vertex was associated with a random number of values (between 2 and 6, 2 and 8, and 2 and 10). The results are summarized in the table below. The two algorithms, A1 and MGA, output loop cutsets of the same size in 55% of the graphs and when the algorithms disagreed, then in 81% of these graphs MGA performed better than A1. The ratio obtained between the number of instances of the algorithms solution and a minimum solution was 1.22 for MGA and 1.44 for A1 (using the 300 graphs in the table below for which the number of vertices is 15 and number of edges 25).

| $|V|$ | $|E|$ | values | A1 | MGA | Eq. |
|---|---|---|---|---|---|
| 15 | 25 | 2–6 | 1 | 17 | 82 |
| 15 | 25 | 2–8 | 2 | 17 | 81 |
| 15 | 25 | 2–10 | 2 | 19 | 79 |
| 55 | 105 | 2–6 | 13 | 58 | 29 |
| 55 | 105 | 2–8 | 17 | 51 | 32 |
| 55 | 105 | 2–10 | 15 | 55 | 30 |
|  |  |  | 50 | 217 | 333 |

To repeat this experiment with A2 required us to make a small change in A2 because it is not designed to run with vertices having different number of values. We adopted the approach of A1 which selects vertices (with at most one parent) according to their degree and if there are several candidates the one with the least number of values is selected for the loop cutset. Combining this idea with the A2 algorithm defines an algorithm we call the weighted A2 algorithm. The results obtained were that MGA performed better than WA2 in 175 of the 224 graphs in which the algorithms disagreed (out of 600). The ratio obtained between the number of instances of the algorithms' solution and a minimum solution was 1.22 for MGA and 1.33 for WA2.

## Remark.

While this work was at its final stages of preparation we became aware of a different method for the WVFS problem that achieves a performance ratio of 2 [Be94]. A quick examination of our own work in light of this information revealed that our method also achieves a performance ratio of 2.